\documentclass{article}

\usepackage{microtype}
\usepackage{graphicx}
\usepackage{subcaption}
\usepackage{booktabs} %
\usepackage{multirow}
\usepackage[svgnames]{xcolor}
\usepackage[normalem]{ulem}
\usepackage[justification=centering]{caption}
\usepackage{tablefootnote}

\definecolor{red}{RGB}{215,118,47}
\definecolor{blue}{RGB}{100,100,204}
\definecolor{green}{RGB}{37,164,83}

\definecolor{light-gray}{gray}{0.95}
\newcommand{\code}[1]{\colorbox{light-gray}{\small{\texttt{#1}}}}

\newcommand{\mytilde}{\raise.17ex\hbox{$\scriptstyle\mathtt{\sim}$}}

\newcommand{\peptides}{\texttt{Peptides}}
\newcommand{\func}{\texttt{Peptides-func}}

\newcommand{\pcqm}{\texttt{PCQM-Contact}}
\newcommand{\voc}{\texttt{PascalVOC-SP}}

\usepackage{hyperref}

\usepackage[accepted]{icml2023}

\usepackage{amsmath}
\usepackage{amssymb}
\usepackage{mathtools}
\usepackage{amsthm}

\usepackage[capitalize,noabbrev]{cleveref}

\theoremstyle{plain}

\theoremstyle{definition}

\theoremstyle{remark}

\newcommand{\ouracro}{\mathrm{DRew}}
\newcommand{\del}{\tau}
\newcommand{\rate}{\nu} %
\newcommand{\R}{\mathbb{R}}

\usepackage[textsize=tiny]{todonotes}

\icmltitlerunning{DRew: Dynamically Rewired Message Passing with Delay}

\begin{document}

\twocolumn[
\icmltitle{DRew: Dynamically Rewired Message Passing with Delay}

\icmlsetsymbol{equal}{*}

\begin{icmlauthorlist}
\icmlauthor{Benjamin Gutteridge}{yyy}
\icmlauthor{Xiaowen Dong}{yyy}
\icmlauthor{Michael Bronstein}{xxx}
\icmlauthor{Francesco Di Giovanni}{sch,usi}

\end{icmlauthorlist}

\icmlaffiliation{yyy}{Department of Engineering Science, University of Oxford}
\icmlaffiliation{xxx}{Department of Computer Science, University of Oxford}
\icmlaffiliation{sch}{Department of Computer Science and Technology, University of Cambridge}
\icmlaffiliation{usi}{Faculty of Informatics, University of Lugano}

\icmlcorrespondingauthor{Benjamin Gutteridge}{beng@robots.ox.ac.uk}

\icmlkeywords{Machine Learning, ICML, Message-Passing Neural Networks, Graph Machine Learning, Geometric Deep Learning, Graph Rewiring}

\vskip 0.3in
]

\printAffiliationsAndNotice{} %

\begin{abstract}
Message passing neural networks (MPNNs) have been shown to suffer from the phenomenon of {\em over-squashing} that causes poor performance for tasks relying on long-range interactions.
This can be largely attributed to message passing only occurring locally, over a node's immediate neighbours.
Rewiring approaches attempting to make graphs `more connected', and supposedly better suited to long-range tasks,
often lose the inductive bias provided by distance on the graph since they make distant nodes communicate {\em instantly} at {\em every} layer.
In this paper we propose a framework, applicable to any MPNN architecture, that performs a \emph{layer-dependent rewiring} to ensure \emph{gradual} densification of the graph. We also propose a \emph{delay} mechanism that permits skip connections between nodes depending on the layer {\em and} their mutual distance.
We validate our approach on several long-range tasks and show that it outperforms graph Transformers and multi-hop MPNNs.
\end{abstract}

\begin{figure*}[tb]
     \centering
     \begin{subfigure}[b]{0.33\textwidth}
         \centering
         \includegraphics[width=\textwidth]{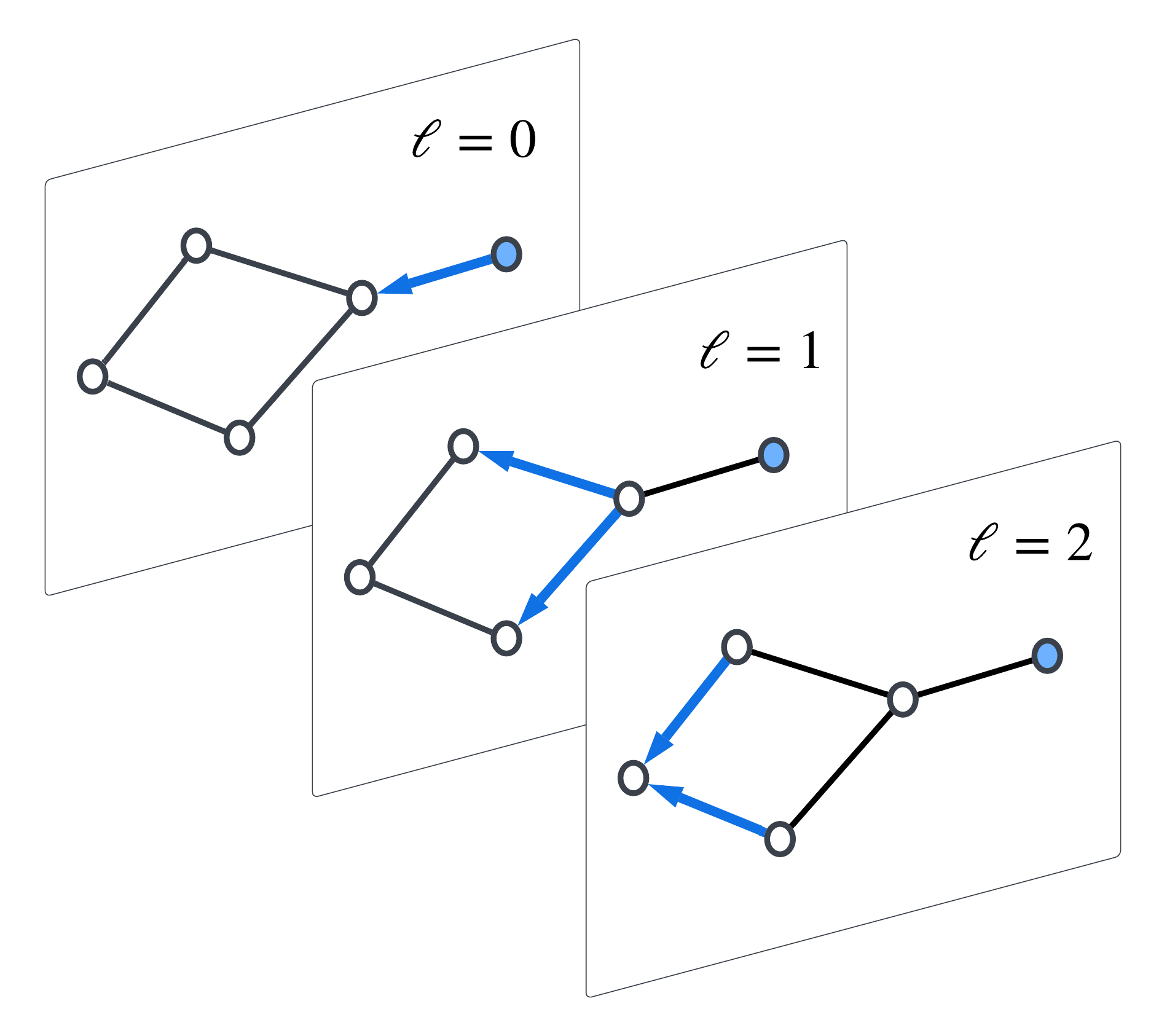}
         \caption{Classical MPNN}
         \label{Fig:mpnn}
     \end{subfigure}
     \hfill
     \begin{subfigure}[b]{0.33\textwidth}
         \centering
         \includegraphics[width=\textwidth]{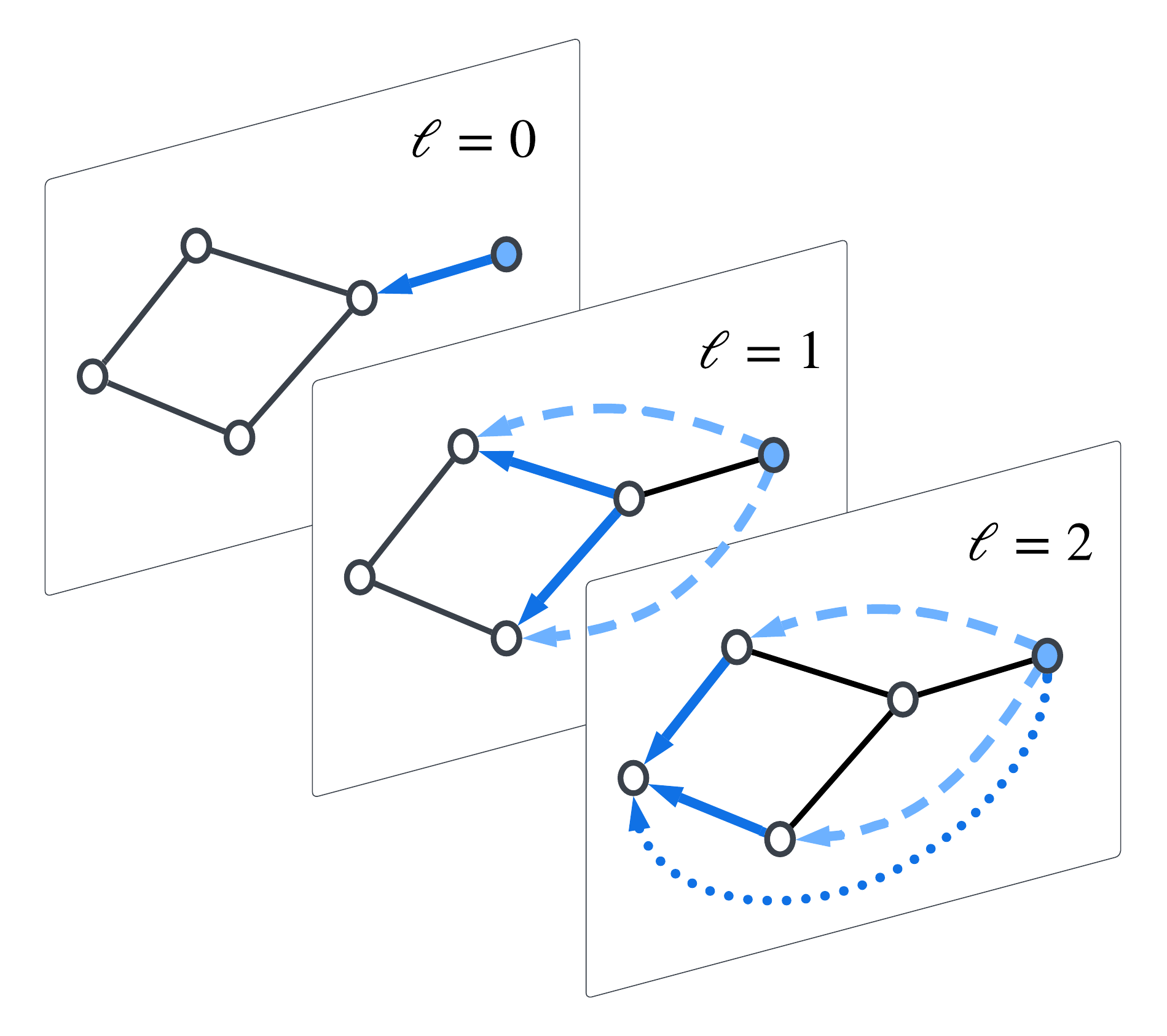}
         \caption{$\ouracro$}
         \label{Fig:non-delay_mpnn}
     \end{subfigure}
     \hfill
     \begin{subfigure}[b]{0.33\textwidth}
         \centering
         \includegraphics[width=\textwidth]{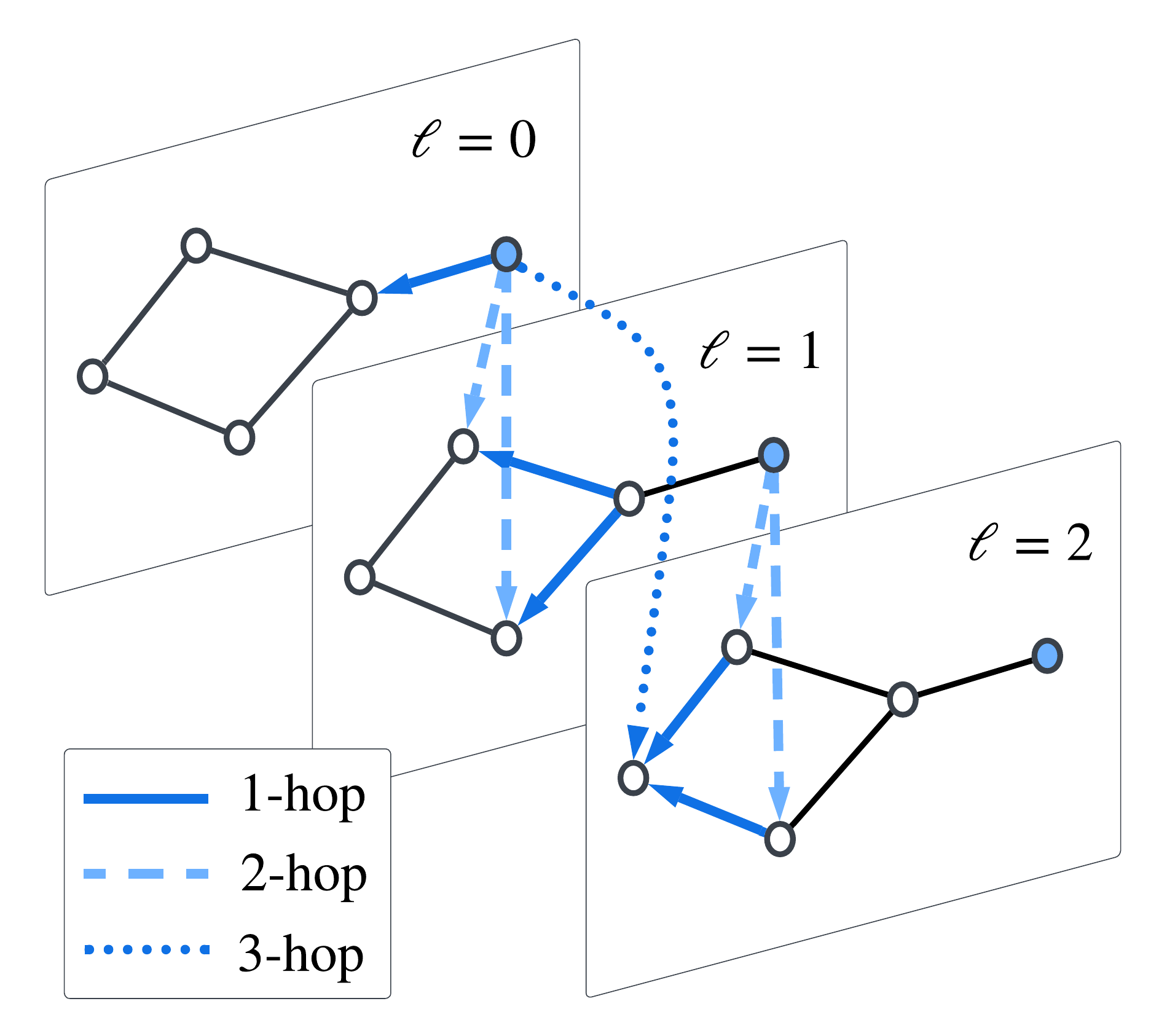}
         \caption{$\rate\ouracro$}
         \label{fig:five over x}
     \end{subfigure}
        \caption{Illustration of the graph across three layers $\ell\in\{0,1,2\}$ for (a) a classical MPNN, (b) $\ouracro$ and (c) $\rate\ouracro$. We choose a source node (coloured blue) on which to focus and demonstrate information flow from this node at each layer.
        We use arrows to denote direction of information transfer and specify hop-connection distance.
        In the classicical MPNN setting, at every layer information only travels from a node to its immediate neighbours. In $\ouracro$, the graph changes based on the layer, with newly added edges connecting nodes at distance $r$ from layer $r-1$ onward. Finally, in $\rate\ouracro$, we also introduce a delay mechanism %
        equivalent to skip-connections between {\em different} nodes based on their mutual distance (see \Cref{subsec:rewiring}).}
        \label{Fig:delay_mpnn}
\end{figure*}

\section{Introduction}
\label{introduction}
Graph Neural Networks (GNNs) \citep{sperduti1994encoding, gori2005new, scarselli2008graph, bruna2013spectral}, deep learning architectures that operate on graph-structured data, are
significantly represented
by the message-passing paradigm \citep{gilmer2017neural}, in which layers  consisting of a local neighbourhood aggregation are stacked to form Message Passing Neural Networks (MPNNs). The most commonly used MPNNs (henceforth referred to as `classical'), perform only {\em local} aggregation, with information being shared at each layer only between nodes that are immediate neighbours (i.e.,  directly connected by an edge). Accordingly, for
nodes that are
distant from one another
to share information, that information must `flow' through the graph at a rate of one edge per layer, necessitating appropriately deep networks when such `long-range interactions' are required for solving the task at hand \citep{barcelo2019logical}. Unfortunately, this often leads to poor model performance, as deep MPNNs are particularly prone to the phenomena of \emph{over-squashing} \citep{alon2020bottleneck,topping2021understanding, di2023over} and \emph{over-smoothing} \citep{nt2019revisiting,Oono2019}.

In this paper, we introduce \textbf{\underline{D}ynamically \underline{Rew}ired Message Passing} ($\ouracro$), a novel framework for layer-dependent, multi-hop message passing that takes a principled approach to information flow, is robust to over-squashing, and can be applied to any MPNN %
for deep learning on graphs.

{\bf Contributions.} %
First, we formalize $\ouracro$, a new framework  of aggregating information over distant nodes that goes beyond the limitations of classical MPNNs, but respects the inductive bias provided by the graph: nodes that are {\em closer} should interact {\em earlier} in the architecture.  %
Second, we introduce the concept of \textbf{delay} for message passing, controlled by a tunable parameter $\rate$, and
generalize $\ouracro$ to
account for delay in order to alleviate issues such as over-smoothing arising from deep MPNN-architectures; we call this framework $\rate\ouracro$.
\footnote{Pronounced `Andrew'.}
Third, we present a theoretical analysis that proves that our proposed frameworks can mitigate over-squashing. Lastly, we experimentally evaluate %
our framework
on both synthetic and real-world datasets.\footnote{\url{https://github.com/BenGutteridge/DRew}} %
Our experiments demonstrate the robustness of $\ouracro$ and the effectiveness of delayed propagation when applied to deep MPNN architectures or long-range tasks.

\section{Message Passing Neural Networks}\label{sec:mpnn}

In this section we introduce the class of Message Passing Neural Networks and discuss some of its main limitations.
We first review some important notions concerning graphs.
\subsection{Preliminaries}
Let $G=(V,E)$ be a graph consisting of nodes $V$ and edges $E$. We assume that $G$ is undirected and connected. The structure of the graph is encoded in the adjacency matrix $\mathbf{A} \subset \R^{n\times n}$, with number of nodes $n=|V|$. The simplest quantity measuring the connectivity of a node is the {\em degree}, which can be computed as $d_i = \sum_{j}A_{ij}$, for $i\in V$. The notion of `locality' in $G$ is induced by the {\em shortest walk} (or {\em geodesic}) {\em distance} $d_G : V\times V \rightarrow \R_{\geq 0}$, which assigns the length of the minimal walk connecting any given pair of nodes. If we fix a node $i$, the distance allows us to partition the graph into level sets of $d_{G}(i,\cdot)$ which we refer to as $k$-{\em hop} (or $k$-{\em neighbourhood}) and denote by
\begin{equation}\label{eq:k-hop}
    \mathcal{N}_{k}(i) := \{ j \in V : d_{G}(i,j) = k\}.\notag
\end{equation}
\noindent %
$\mathcal{N}_{1}(i)$ is the set of
immediate (or 1-hop) neighbours of node $i$.
We stress that %
in our notations, the $k$-hop of a node $i$ represents the nodes at distance {\em exactly} $k$ --- %
a subset of the nodes that can be reached by a walk of length $k$. %

\paragraph{The MPNN class.} Consider a graph $G$ with node features $\{ h_i \in \R^{d}, i \in V\}$ and assume we are interested in predicting a quantity (or label) $y_{G_{i}}$. Typically a GNN processes both the topological data $G$ and the feature information $\mathbf{H}\in\R^{n\times d}$ via a sequence of layers, before applying a readout map to output a final prediction $h_{G}$. The most studied GNN paradigm is MPNNs \cite{gilmer2017neural}, where the layer update is given by
\begin{equation}
\begin{split}
\label{Eq:static_mpnn}
a_{i}^{(\ell)} &= \mathrm{AGG}^{(\ell)}\left(\{h_j^{(\ell)} : j \in \mathcal{N}_1(i)\}\right),\\
h^{(\ell+1)}_i &= \mathrm{UP}^{(\ell)}\left( h_i^{(\ell)}, a_i^{(\ell)}\right),
\end{split}
\end{equation}
\noindent for learnable update and aggregation maps $\mathrm{UP}$ and $\mathrm{AGG}$. %
While the choice of the maps $\mathrm{UP}$ and $\mathrm{AGG}$ may change across specific architectures \citep{bresson2017residual,hamilton2017inductive,kipf2016semi, velivckovic2017graph}, in all MPNNs messages travel from one node to its 1-hop neighbours {\em at each layer}. Accordingly, %
for a node $i$ to exchange information with node $j\in\mathcal{N}_k(i)$, we need to stack at least $k$ layers. %
In \Cref{sec:our_framework}, we %
discuss how the interaction between two node representations should, in fact, change based on both their mutual distance {\em and} their state in time (i.e. the layer of the network). We argue that it is important \textbf{not simply {\em how}} two node states interact with each other, \textbf{but also \textbf{\em when}} that happens. %

\subsection{Long-range dependencies and network depth}

A task exhibits {\em long-range interactions} if, to be solved, there exists some node $i$ whose representation needs to account for the information contained at
a node
$j$ %
with $d_G(i,j)\gg 1$
\citep{dwivedi2022long}.
MPNNs rely on 1-hop message propagation, so to capture such non-local interactions, multiple layers must be stacked; however, this leads to undesirable phenomena with increasing network depth. We focus on one such problem, known as over-squashing, below. %

\paragraph{Over-squashing.}
\label{sec:oversquashing}
In a graph, the number of nodes in the receptive field of a node $i$ often expands exponentially with hop distance $k$. Accordingly, for $i$ to exchange information with its $k$-hop neighbours, an exponential volume of messages must pass through fixed-size node representations, which may ultimately lead to a loss of information \citep{alon2020bottleneck}. This problem is known as {\em over-squashing}, and has been characterized  via sensitivity analysis \citep{topping2021understanding}.
Methods to address the over-squashing problem typically resort to some form of {\em graph rewiring}, in the sense that the graph used for message passing is (partly) decoupled from the input one. %
A `local' form of graph rewiring consists in  aggregating over multiple hops at each layer layer \citep{abu2019mixhop, abu2020n, zhang2021nested, abboud2022shortest}. A `global' form of graph rewiring is taken to the extreme in graph Transformers \cite{ying2021transformers,kreuzer2021rethinking,rampavsek2022recipe}, which
replace the input graph with a complete graph where every pair of nodes is connected by an attention-weighted edge.
Transformers, however, are computationally expensive and tend to throw away information afforded by the graph topology. Since all nodes can interact in a single layer, any notion of locality induced by distance $d_G$ is discarded and must be rediscovered implicitly via positional and structural encoding.

\paragraph{Over-smoothing and other phenomena.} The use of deep MPNNs gives rise to other issues beyond over-squashing. A well-known problem is {\em over-smoothing}, where, in the limit of many layers, features become indistinguishable %
\citep{nt2019revisiting, Oono2019}. While over-smoothing is now fairly understood and has been formally characterized in recent works \citep{bodnar2022neural,cai2020note,di2022graph,rusch2022gradient}, it is unclear whether the often observed degradation in performance with increasing depth is mainly caused by over-smoothing,
over-squashing, or more classical vanishing gradient problem \citep{di2023over}.
It is hence generally desirable to propose frameworks that are not just robust to depth, but can actually adapt to the underlying task, either by `fast' exploration of the graph in fewer layers or by
`slow' aggregation through multiple layers. {\em We introduce a new framework for message-passing that can accomplish this thanks to two principles: (i) dynamically rewired message passing and (ii) a delay mechanism.}

\section{Dynamically Rewired MPNNs}\label{sec:our_framework}
In this section we introduce our framework to handle the aggregation of messages in MPNNs. We discuss how MPNNs present a `static' way of collecting messages at each layer which is ultimately responsible for over-squashing. By removing such static inductive bias, we unlock a physics-inspired way for MPNNs to exchange information that is more suited to handle long-range interactions.

\paragraph{Information flow in MPNNs.} Consider two nodes $i,j \in V$ at distance $r$. In a classic MPNN, these two nodes start interacting
at layer $r$, meaning that
\begin{equation}
\label{eq:first_layer_distance}
\min \Big\{ \ell : \frac{\partial h_{i}^{(\ell)}}{\partial h_j^{(0)}} \neq 0\Big\} \geq d_{G}(i,j).
\end{equation}

\noindent In fact, since the aggregation at each layer is computed using the same graph $G$, one can bound such interaction with powers of the adjacency $\mathbf{A}$ as used in \citet{topping2021understanding}
\begin{equation}\label{eq:jacobian_bound_classic}
    \left\vert \frac{\partial h_{i}^{(r)}}{\partial h_j^{(0)}} \right\vert \leq c\, (\mathbf{A}^{r})_{ij},
\end{equation}
\noindent with constant $c$ depending only on the Lipschitz-regularity of the MPNN and independent of the graph topology. We see that communication between $i$ and $j$ must be filtered by intermediate nodes that are traversed along each path connecting $i$ to $j$. This, in a nutshell, is the reason behind over-squashing; indeed, the bound in Eq.~\eqref{eq:jacobian_bound_classic} may decay exponentially with the distance $r$ whenever $\mathbf{A}$ is degree-normalized. By the same argument, %
in an MPNN two nodes at distance $r$ always interact with a latency or {\em delay of exactly} $r$, meaning that for any intermediate state $\ell_0$ we have
\begin{equation}
\min \Big\{ \ell : \frac{\partial h_{i}^{(\ell)}}{\partial h_j^{(\ell_0)}} \neq 0\Big\} \geq \ell_0 + d_{G}(i,j),
\end{equation}
\noindent and similar Jacobian bounds apply in this case. Accordingly, in a classic MPNN we have two problems:
\begin{itemize}
\item[(i)] Distant nodes can only communicate by exchanging information with their neighbours.
\item[(ii)] Distant nodes always interact with a fixed delay given by their distance.
\end{itemize}

\paragraph{Information flow in multi-hop MPNNs.} The first issue can, in principle, be easily addressed by {\em rewiring} the graph via a process where %
any pair of nodes within a certain threshold are connected via an edge, and which can generally be given its own encoding or weight \citep{abboud2022shortest, bruel2022rewiring, rampavsek2022recipe}.  In this way, distant nodes can now exchange information {\em directly}; this avoids iterating messages through powers of the adjacency and hence mitigates over-squashing by reducing the exponent in Eq.~\eqref{eq:jacobian_bound_classic}. However, this process brings about two phenomena which could lead to undesirable effects: (i) the computational graph is much denser --- with implications for efficiency --- and (ii) most of the inductive bias afforded by the graph distance information is thrown away, given that nodes $i,j$ at distance $r$ are now able to interact {\em at each layer} of the architecture, without any form of latency. In particular, this {\bf static} rewiring,
where the computational graph is densely `filled' from the first MPNN layer, prevents messages from being sent \emph{first} among nodes that are closer together in the input graph.

\subsection{A new framework: $(\rate)\ouracro$ message passing}

\paragraph{Dynamic rewiring.} We start by addressing the limitation of MPNNs that nodes can only communicate through intermediate neighbours. %
To motivate our framework, take two nodes $i,j\in V$ at distance $r > 1$. For classical MPNNs, we must wait for $r$ layers (i.e. time units {\em with respect to the architecture}) before $i$ and $j$ can interact with each other. As argued above, this preserves information about locality and distance induced by the input graph, since nodes that are {\em closer} communicate {\em earlier}; however, since the two nodes have waited `long enough', we argue that they should interact directly without necessarily relaying messages to their neighbours first. Accordingly, given update and aggregation functions as per the MPNN paradigm in Eq.~\eqref{Eq:static_mpnn}, we define the update in a \textbf{Dynamically Rewired (}$\text{\textbf{DRew}}$-\textbf{)MPNN} by:
\begin{align}
\label{Eq:dynamic_mpnn}
a_{i,k}^{(\ell)} &= \mathrm{AGG}^{(\ell)}_{k}\Big(\{h_j^{(\ell)} : j \in \mathcal{N}_{k}(i)\}\Big), 1\leq k \leq \ell+1 \notag \\
h^{(\ell+1)}_i &= \mathrm{UP}_k^{(\ell)}\Big( h_i^{(\ell)}, a_{i,1}^{(\ell)}, \ldots, a_{i,\ell+1}^{(\ell)}\Big).
\end{align}
\noindent Some important comments: first, if $\mathrm{AGG}_k = I$ for each $k > 1$, this reduces to the classical MPNN setting. Second, unlike augmented MPNNs, %
the sets over which we compute aggregation {\em differ depending on the layer}, with the hop $\mathcal{N}_k(i)$ only being added from the $k$-th layer on. So, while this framework shares similarities with other multi-hop MPNN architectures like \citet{abboud2022shortest}, it features a novel mechanism: dynamic rewiring of the graph at each layer $\ell$ to include aggregation from each $k$-hop within distance $\ell+1$. For example, at the first layer, $\ell=0$, our $\ouracro$ layer is identical to the base MPNN represented by the choice of $\mathrm{UP}$ and $\mathrm{AGG}$, but at each subsequent layer the receptive field of node $i$ expands by 1 hop. This allows distant nodes to exchange information without intermediate steps, hence solving one of the problems of the MPNN paradigm, but also preserving the inductive bias afforded by the topology since the graph is filled gradually according to distance rather than treating each layer in the same way. %

{\em $\ouracro$-MPNN explores the middle ground between classical MPNNs
and methods like graph Transformers that consider all pairwise interactions at once}. %

\paragraph{The delay mechanism.}
Next, we %
generalize $\ouracro$-MPNN to also account for  whether nodes should interact with fixed (if any) delay. Currently, we have two opposing scenarios: in MPNNs, nodes interact with a constant delay given by their %
distance -- leading to the same lag of information -- while in $\ouracro$, nodes interact only from a certain depth of the architecture, but without any delay. For $\ouracro$, two nodes $i,j$ at distance $r$ communicate directly after $r$ layers, since information has now been able to travel from $j$ to $i$.
But what if we consider the state of $j$ as it was when the information `left' to flow towards $i$?
We %
account for an extra degree of freedom $\del$ representing the \textbf{delay} of messages exchanged among nodes at distance $r$ in $\ouracro$. %
If $\del = 0$, then nodes at distance $k$ interact at the $k$-th layer without delay,
i.e {\em instantly} as per Eq.~\eqref{Eq:dynamic_mpnn},
otherwise node $i$ `sees' the state of $j$ at the $k$-th layer but \emph{delayed} by $\del := k - \rate$, for some $\rate$. We formalize this by introducing
$\del_\nu(k) = \max(0, k-\rate)$ and generalize $\ouracro$ as %
\begin{align}
\label{Eq:dynamic_mpnn_delay}
a_{i,k}^{(\ell)} &= \mathrm{AGG}^{(\ell)}_{k}\Big(\{h_j^{\left(\ell-\del_\rate(k)\right)} : j \in \mathcal{N}_{k}(i)\}\Big), 1\leq k \leq \ell+1 \notag \\
h^{(\ell+1)}_i &= \mathrm{UP}_k^{(\ell)}\Big( h_i^{(\ell)}, a_{i,1}^{(\ell)}, \ldots, a_{i,\ell+1}^{(\ell)}\Big).
\end{align}
\noindent If there is no delay, i.e. $\rate = \infty$,
then we recover Eq.~\eqref{Eq:dynamic_mpnn}.
The opposite case is given by $\rate = 1$, so that at layer $\ell$ and for any $j$ at distance $k$, node $i$ receives `delayed' representation $h_j^{(\ell - \tau_1(k))}$, i.e. the state of $j$ as it was $k-1$ layers ago.
From now on, we refer to Eq.~\eqref{Eq:dynamic_mpnn_delay} as $\rate\ouracro$. We also note that in our experiments we treat $\rate$ as a tunable hyperparameter. %

{\em Delay allows for expressive control of information flow. No delay means that messages travel faster, with distant nodes interacting instantly once an edge is added; conversely, the more delay, the slower the information flow, with distant nodes accessing \textbf{past} states when an edge is added.}

Our framework \emph{generalizes any MPNN} since it acts on the computational graph (which nodes exchange information and {\em when}) and does not govern
the architecture (see \Cref{subsec:rewiring}). We describe three instances of $\rate\ouracro$ below.

\subsection{Instances of our framework}
\label{sec:instances}
In this section we provide examples for the $\rate\ouracro$-MPNN template in Eq.~\eqref{Eq:dynamic_mpnn_delay} for three classical MPNNs: GCN \citep{kipf2016semi}, GIN \cite{xu2018powerful} and GatedGCN \cite{bresson2017residual}. We will use these variants for our experiments in Section \ref{sec:experiments}. For $\rate\ouracro$-GCN, we write the layer-update as
\begin{equation}
\label{Eq:drewGCN}
   h^{(\ell+1)}_i = h^{(\ell)}_i + \sigma \left( \sum_{k = 1}^{\ell+1} \sum_{j \in \mathcal{N}_k(i)} \mathbf{W}_{k}^{(\ell)} \gamma^{k}_{ij} h_{j}^{\left(\ell-\del_\nu(k)\right)} \right),
\end{equation}
\noindent where $\sigma$ is a pointwise nonlinearity, $\mathbf{W}_k^{(\ell)}$ are learnable channel-mixing matrices for the convolution at layer $\ell$ on the $k$-neighbourhood, and $\boldsymbol{\Gamma}^k \subset \R^{n \times n}$ are matrices with elements
\begin{equation}\label{eq:gamma}
\gamma^k_{ij} =
            \begin{cases}
		\frac{1}{\sqrt{d_i d_j}}, & \text{if } d_G(i,j) = k\\
            0, & \text{otherwise}.
		 \end{cases}
\end{equation}
 \noindent We note again that if $j\in\mathcal{N}_{k}(i)$, then $i,j$ only communicate
 from layer $k$ onward, while $\rate$ determines the communication delay. The choice of degree normalization for $\boldsymbol{\Gamma}^k$ is to provide a consistent normalization for all terms.

We define $\rate\ouracro$-GIN in a similar fashion, as per Eq.~\eqref{Eq:dynamic_mpnn_delay}; the layer update used below is inspired by \citet{brockschmidt2020gnn} and \citet{abboud2022shortest}:
\begin{gather}
\begin{split}
\label{Eq:r*GIN}
    h_{i}^{(\ell+1)} = &\; (1+\epsilon)\mathrm{MLP}_{s}^{(\ell)}(h_{i}^{(\ell)}) \\
                            &+ \sum_{k = 1}^{\ell+1}\,\sum_{j \in \mathcal{N}_k(i)}\mathrm{MLP}_{k}^{(\ell)}(h^{\left(\ell - \del_\rate(k)\right)}_{j}),
\end{split}
\end{gather}
where an $\text{MLP}$ is one or more linear layers separated by ReLU activation and $\epsilon$ is a weight parameter. $\text{MLP}_s^{(\ell)}$ is the self-loop (or residual) aggregation while $\text{MLP}_k^{(\ell)}$ operates on the $k$-neighbourhood at layer $\ell$.

Lastly, we define $\rate\ouracro$-GatedGCN as follows:
\begin{gather}
\label{eq:drew-gatedgcn}
\begin{split}
    h^{(\ell+1)}_i &= \mathbf{W}_1^{(\ell)} h^{(\ell)}_i + \sum_{k = 1}^{\ell+1} \sum_{j \in \mathcal{N}_k(i)} \eta^k_{i,j} \odot \mathbf{W}_2^{(\ell)} h_j^{(\ell - \tau_\nu (k))}, \\
    \eta_{i,j}^k &= \frac{\hat{\eta}_{i,j}^k}{\sum_{j \in \mathcal{N}_k(i)} ( \hat{\eta}_{i,j}^k ) + \epsilon}, \\
    \hat{\eta}_{i,j}^k &= \sigma \left ( \mathbf{W}_3^{(\ell)} h_i^{(\ell)} + \mathbf{W}_4^{(\ell)} h_j^{(\ell - \tau_\nu(k))}\right ),
\end{split}
\end{gather}
where $\sigma$ is the sigmoid function, $\odot$ is the element-wise product, $\epsilon$ is a small fixed constant for numerical stability and $\mathbf{W}_1^{\ell}, \mathbf{W}_2^{\ell}, \mathbf{W}_3^{\ell}, \mathbf{W}_4^{\ell}$ are learned channel-mixing matrices. We note that in Eq.~\eqref{eq:drew-gatedgcn}, unlike Eq.~\eqref{Eq:drewGCN} and Eq.~\eqref{Eq:r*GIN}, weight matrices are shared between $k$-hop neighbourhoods. We do this because $k$-neighbourhood weight sharing achieves comparably strong results with non-weight-shared $\ouracro$-GatedGCN (see Section \ref{sec:experiments}) while maintaining a lower parameter count, whereas we see performance drops when using weight sharing for $\ouracro$-GCN and -GIN. This can be explained by the edge gates $\eta^k_{i,j}$ serving as a soft-attention mechanism \cite{dwivedi2020benchmarking} not present in GCN and GIN, affording shared weights more flexibility to model relationships between nodes at varying hop distances.

\subsection{The graph-rewiring perspective: $\rate\ouracro$ as distance-aware skip-connections}\label{subsec:rewiring}
We conclude this section by providing an alternative explanation of $\rate\ouracro$ from a graph-rewiring perspective. Given an underlying MPNN, %
we study how information travels in the graph at each layer.
Referring to \Cref{Fig:delay_mpnn} to illustrate our explanation,
we say that messages travel {\bf horizontally} when they move inside a layer (slice), and that they move {\bf vertically} when they travel across different layers (slices). In a classical MPNN, the graph adopted at each layer coincides with the input graph $G$; information can only travel horizontally from a node to its 1-hop neighbours. In the $\ouracro$ setting --- which we recall to be the version of $\rate\ouracro$ without delay (i.e. $\rate = \infty$) --- the graph changes {\em depending on the layer}: this amounts to a dynamic rewiring where $G$ is replaced with a sequence $\{\mathcal{R}_k(G)\}$, where at each layer $\ell$ we add edges between any node $i$ and $\mathcal{N}_{\ell+1}(i)$. Messages only travel horizontally as before, but the  graph is {\em progressively filled} with each layer.
Finally, in the delayed version of Eq.~\eqref{Eq:dynamic_mpnn_delay}, messages can travel both horizontally {\em and} vertically, meaning that we are %
also `rewiring' the graph along the time (layer) axis.
Residual connections can also be thought of as allowing information to move `vertically', though such connections are only made between the \emph{same} node $i$ at \emph{different} layers; typically $\ell, \ell+1$. From this perspective, the $\rate\ouracro$ framework {\bf is equivalent to adding geometric skip connections} among \emph{different} nodes based on their distance. This is a powerful mechanism that combines skip-connections, a key tool in architecture design, with metric information provided by the geometry of the data; in this case the distances between vertices in a graph $G$.

\begin{table*}[ht]
\centering
\caption{\label{tab:mpnn_vs_drew_all_lrgb}Classical MPNN benchmarks vs their $\ouracro$ variants (without positional encoding) across four LRGB tasks: (from left to right) graph classification, graph regression, link prediction and node classification. All results are for the given metric on test data.}
\begin{tabular}{@{}rcccc@{}}
\toprule
\multirow{2}{*}{\textbf{Model}} & \texttt{Peptides-func} & \texttt{Peptides-struct} & \texttt{PCQM-Contact}  & \texttt{PascalVOC-SP}  \\
                                & AP $\uparrow$          & MAE $\downarrow$         & MRR $\uparrow$         & F1 $\uparrow$          \\ \midrule
GCN                             & 0.5930\small±0.0023          & 0.3496\small±0.0013            & 0.3234\small±0.0006          & 0.1268\small±0.0060          \\
+$\ouracro$                           & \textbf{0.6996\small±0.0076} & \textbf{0.2781\small±0.0028}   & \textbf{0.3444\small±0.0017} & \textbf{0.1848\small±0.0107} \\ \midrule
GINE                            & 0.5498\small±0.0079          & 0.3547\small±0.0045            & 0.3180\small±0.0027          & 0.1265\small±0.0076          \\
+$\ouracro$                           & \textbf{0.6940\small±0.0074} & \textbf{0.2882\small±0.0025}   & \textbf{0.3300\small±0.0007} & \textbf{0.2719\small±0.0043} \\ \midrule
GatedGCN                        & 0.5864\small±0.0077          & 0.3420\small±0.0013            & 0.3218\small±0.0011          & 0.2873\small±0.0219          \\
+$\ouracro$                           & \textbf{0.6733\small±0.0094} & \textbf{0.2699\small±0.0018}   & \textbf{0.3293\small±0.0005} & \textbf{0.3214\small±0.0021} \\ \bottomrule
\end{tabular}
\end{table*}

\section{Why $\mathbf{\nu}\text{\textbf{DRew}}$ Improves Information Processing}
\label{sec:theory}
\paragraph{Mitigating over-squashing.} In this section we discuss why the $\rate\ouracro$ framework mitigates over-squashing and is hence more suited to handle long-range interactions in a graph. We focus on the case of maximal delay $\rate = 1$. %
We also restrict our discussion to Eq.~\eqref{Eq:drewGCN}: $\rate\ouracro$-GCN, though the conclusion extends easily to any $\rate\ouracro$-MPNN. Consider nodes $i,j \in V$ at distance $r$. For a traditional MPNN, $i,j$ first exchange information at layer $r$, meaning that Eq.~\eqref{eq:first_layer_distance} is satisfied; however, a crucial difference from the MPNN paradigm is given by the addition of distance-aware skip connections between $i,j$ as in Eq.~\eqref{Eq:dynamic_mpnn_delay}. One can extend the approach from \citet{topping2021understanding} and derive
\begin{equation*}
    \left\vert \frac{\partial h_{i}^{(r)}}{\partial h_j^{(0)}}\right\vert \leq C\Big(\sum_{k_1 + \dots + k_\ell = r}\Big(\prod_{k_1,\ldots,k_\ell}(\gamma^k)_{ij}\Big)\Big),
\end{equation*}
\noindent recalling that matrices $\boldsymbol{\Gamma}^k$ are defined in Eq.~\eqref{eq:gamma}. We see how, differently from the standard MPNN formalism, nodes at distance $r$ can now interact via products of message-passing matrices containing {\em fewer} than $r$ factors. In fact, the right-hand side also accounts for a direct interaction between $i,j$ via the matrix $\boldsymbol{\Gamma}^{r}$.
\citet{topping2021understanding} showed that over-squashing arises precisely due to the entries $ij$ of $\mathbf{A}^{r}$ decaying to zero exponentially with the distance, $r$, for (normalized) message-passing matrices $\mathbf{A}$; on the other hand, using matrices like $\boldsymbol{\Gamma}^r$, which are not powers of the same adjacency matrix, mitigates over-squashing.

\paragraph{Interpreting delay as local smoothing.}\label{subsec:over-smoothing_analysis}
We comment here on a slightly different perspective from which to understand the role of delay in $\rate\ouracro$. As usual, we consider nodes $i,j$ at distance $r$. In our framework, node $i$ starts collecting messages from $j$ starting from the $r$-th layer. A larger delay (i.e. smaller value of $\rate$), means that $i$ aggregates the features from $j$ before they are (significantly) `smoothed' by repeated message passing.
Conversely, a smaller delay (i.e. larger value of $\rate$), implies that when $i$ communicates with $j$, it also leverages the structure around $j$ which has been encoded in the representation of $j$ via the earlier layers. Therefore, we see how beyond mitigating over-squashing, the delay offers an extra degree of freedom to our framework, which can be used to {\em adapt to the underlying task and how quickly the graph topological information needs to be mixed across different regions}.

\paragraph{Expressivity.}
As multi-hop aggregations used in $\rate\ouracro$ are based on shortest path distances, they are able to distinguish any pair of graphs distinguished by the shortest path kernel \citep{abboud2022shortest,borgwardt2005shortest}. Shortest path can distinguish disconnected graphs,
a task at which 1-WL \citep{weisfeiler1968reduction}, which bounds the expressiveness of classical MPNNs \citep{xu2018powerful}, fails. We can therefore state that, at minimum, $(\rate)\ouracro$ is more expressive than 1-WL and, therefore, classical MPNNs. We leave more detailed expressivity analysis for future work.

\section{Empirical Analysis}
\label{sec:experiments}
In this section we focus on two strengths of our model. First, we validate \textbf{performance} in comparison with benchmark models, including vanilla and multi-hop MPNNs and graph Transformers, over five real-world tasks %
spanning graph-, node- and edge-level tasks. Second, we validate the \textbf{robustness} of $\rate\ouracro$ for long-range-dependent tasks and increased-depth architectures, using a synthetic task and a real-world molecular dataset. 

\vspace{-1mm}

\paragraph{Parameter scaling.} We note here that many of the $\ouracro$-MPNNs used in our experiments exhibit parameter scaling of approximately $L^2/2$ for network depth $L$, whereas MPNNs scale with $L$. For fair comparison, a {\em fixed parameter budget} is maintained for all performance experiments across all network depths via suitable adjustment of hidden dimension $d$, for both MPNNs and Transformers -- %
we reserve the exploration of optimal sharing of weights for future work. We discuss space-time complexity in Appendix~\ref{appendix:complexity}.

\begin{table*}[h]
\centering
\caption{Performance of various classical, multi-hop and static rewiring MPNN and graph Transformer benchmarks against $\ouracro$-MPNNs across four LRGB tasks. The \textbf{\textcolor{green}{first-}}, \textbf{\textcolor{red}{second-}} and \textbf{\textcolor{blue}{third-}}best results for each task are colour-coded; models whose performance are within a standard deviation of one another are considered equal.}
\label{tab:all_results}
\begin{tabular}{@{}lcccc@{}}
\toprule
\multirow{2}{*}{\textbf{Model}} & \texttt{Peptides-func}                    & \texttt{Peptides-struct}                  & \texttt{PCQM-Contact}                     & \texttt{PascalVOC-SP}                     \\
                                & AP $\uparrow$                             & MAE $\downarrow$                          & MRR $\uparrow$                            & F1 $\uparrow$                             \\ \midrule
GCN                             & 0.5930\small±0.0023                             & 0.3496\small±0.0013                             & 0.3234\small±0.0006                             & 0.1268\small±0.0060                             \\
GINE                            & 0.5498\small±0.0079                             & 0.3547\small±0.0045                             & 0.3180\small±0.0027                             & 0.1265\small±0.0076                             \\
GatedGCN                        & 0.5864\small±0.0077                             & 0.3420\small±0.0013                             & 0.3218\small±0.0011                             & 0.2873\small±0.0219                             \\
GatedGCN+PE     & 0.6069\small±0.0035                             & 0.3357\small±0.0006                             & 0.3242\small±0.0008                             & 0.2860\small±0.0085                             \\ \midrule
DIGL+MPNN       & 0.6469\small±0.0019                             & 0.3173\small±0.0007                             & 0.1656\small±0.0029                             & 0.2824\small±0.0039                             \\
DIGL+MPNN+LapPE & 0.6830\small±0.0026                             & \textbf{\textcolor{blue}{0.2616\small\small±0.0018}}  & 0.1707\small±0.0021                             & 0.2921\small±0.0038                             \\
MixHop-GCN                      & 0.6592\small±0.0036                             & 0.2921\small±0.0023                             & 0.3183\small±0.0009                             & 0.2506\small±0.0133                             \\
MixHop-GCN+LapPE                & 0.6843\small±0.0049                             & \textbf{\textcolor{blue}{0.2614\small±0.0023}}  & 0.3250\small±0.0010                             & 0.2218\small±0.0174                             \\ \midrule
Transformer+LapPE               & 0.6326\small±0.0126                             & \textbf{\textcolor{red}{0.2529\small±0.0016}}   & 0.3174\small±0.0020                             & 0.2694\small±0.0098                             \\
SAN+LapPE                       & 0.6384\small±0.0121                             & 0.2683\small±0.0043                             & \textbf{\textcolor{blue}{0.3350\small±0.0003}}  & \textbf{\textcolor{blue}{0.3230\small±0.0039}}  \\
GraphGPS+LapPE                  & 0.6535\small±0.0041                             & \textbf{\textcolor{green}{0.2500\small±0.0005}} & 0.3337\small±0.0006                             & \textbf{\textcolor{green}{0.3748\small±0.0109}} \\ \midrule
DRew-GCN                        & \textbf{\textcolor{blue}{0.6996\small±0.0076}}  & 0.2781\small±0.0028                             & \textbf{\textcolor{green}{0.3444\small±0.0017}} & 0.1848\small±0.0107                             \\
DRew-GCN+LapPE                  & \textbf{\textcolor{green}{0.7150\small±0.0044}} & \textbf{\textcolor{red}{0.2536\small±0.0015}}   & \textbf{\textcolor{green}{0.3442\small±0.0006}} & 0.1851\small±0.0092                             \\
DRew-GIN                        & \textbf{\textcolor{blue}{0.6940\small±0.0074}}  & 0.2799\small±0.0016                             & 0.3300\small±0.0007                             & 0.2719\small±0.0043                             \\
DRew-GIN+LapPE                  & \textbf{\textcolor{red}{0.7126\small±0.0045}}   & \textbf{\textcolor{blue}{0.2606\small±0.0014}}  & \textbf{\textcolor{red}{0.3403\small±0.0035}}   & 0.2692\small±0.0059                             \\
DRew-GatedGCN                   & 0.6733\small±0.0094                             & 0.2699\small±0.0018                             & 0.3293\small±0.0005                             & \textbf{\textcolor{blue}{0.3214\small±0.0021}}  \\
DRew-GatedGCN+LapPE             & \textbf{\textcolor{blue}{0.6977\small±0.0026}}  & \textbf{\textcolor{red}{0.2539\small±0.0007}}   & 0.3324\small±0.0014                             & \textbf{\textcolor{red}{0.3314\small±0.0024}}   \\ \bottomrule
\end{tabular}
\end{table*}

\subsection{Long-range graph benchmark}
\label{sec:lrgb}
The Long Range Graph Benchmark (LRGB; \citet{dwivedi2022long}) is a set of GNN benchmarks involving long-range interactions. We provide experiments for three datasets from this benchmark (two molecular property prediction, one image segmentation) spanning the full range of tasks associated with GNNs: graph regression (\texttt{Peptides-func}), graph classification (\texttt{Peptides-struct}), link prediction (\texttt{PCQM-Contact}) and node classification (\texttt{PascalVOC-SP}). The tasks presented by LRGB are characterised as possessing long-range dependencies according to the criteria of (a) graph size (i.e. having a large number of nodes), (b) requiring a long range of interaction, and (c) output sensitivity to global graph structure. We compare our $\ouracro$-MPNN variants from Section \ref{sec:instances} against classical MPNN benchmarks in Table \ref{tab:mpnn_vs_drew_all_lrgb}, and against a range of models in Table \ref{tab:all_results}, including classical and $\ouracro$ MPNN variants,  graph Transformers \cite{dwivedi2022long, rampavsek2022recipe}, a multi-hop baseline (MixHop-GCN; \citet{abu2019mixhop}) and a static graph rewiring benchmark (DIGL; \citet{klicpera2019diffusion}).

\paragraph{Experimental details.}
All experiments are averaged over three runs and were allowed to train for 300 epochs or until convergence. Classical MPNN and graph Transformer results are reproduced from \citet{dwivedi2022long}, except GraphGPS which is reproduced from \citet{rampavsek2022recipe}. $\ouracro$-MPNN, DIGL and MixHopGCN models were trained using similar hyperparameterisations to their classical MPNN counterparts (see Appendix \ref{appendix:experiments}. Some models include positional encoding (PE), either Laplacian (LapPE; \citet{dwivedi2020benchmarking}) or Random Walk (RWSE; \citet{dwivedi2021graph}), as this improves performance and is necessary to induce a notion of locality in Transformers. We provide the performance of the best-case $\rate\ouracro$ model with respect to $\rate \in \{1, \infty\}$ and network depth $L$ for both the PE and non-PE cases. Hyperparameters and other experimental details are available in Appendix \ref{appendix:experiments}.
As in \citet{dwivedi2022long},
\textbf{we use a fixed \mytilde500k parameter budget.}

\paragraph{Discussion.}
As shown in Table \ref{tab:mpnn_vs_drew_all_lrgb}, $\rate\ouracro$-MPNNs substantially outperform their classical counterparts across all four tasks. We particularly emphasise this result for GINE and GatedGCN, as both models utilise edge features, unlike their $\ouracro$ counterparts.
Furthermore, $\ouracro$ outperforms the static rewiring and multi-hop benchmarks in all tasks, and at least one $\ouracro$-MPNN model matches or beats the best `classical' graph Transformer baseline from \citet{dwivedi2022long} \emph{in all four tasks}. GraphGPS \cite{rampavsek2022recipe} outperforms the best $\ouracro$ model in the \texttt{PascalVOC-SP} and and \texttt{Peptides-struct} tasks, but we stress that GraphGPS is a much more sophisticated architecture that \emph{combines dense Transformers with message passing}, and therefore supports our claim that pure global attention throws away important inductive bias afforded by MPNN approaches. Even so, $\ouracro$ still surpasses GraphGPS in the \texttt{Peptides-func} and \texttt{PCQM-Contact} tasks.

\subsection{\texttt{QM9}}
\texttt{QM9} \cite{ramakrishnan2014quantum} is a molecular multi-task graph regression benchmark dataset of \mytilde130,000 graphs with \mytilde18 nodes each and a maximum graph diameter of 10.
We compare $\rate\ouracro$-GIN against a number of benchmark MPNNs and a GIN-based multi-hop MPNN: shortest path network (SPN; \citet{abboud2022shortest}), which is similar to our work in that it uses a multi-hop aggregation based on shortest path distances, but differs crucially in the lack of dynamic, per-layer rewiring or delay. Experimental results for all regression targets are given in Table \ref{tab:qm9}.

\begin{table*}[h]
\caption{Performance of $\rate\ouracro$ compared with MPNN benchmarks on \texttt{QM9}. Scores reported are test MAE, i.e. lower is better.}
\label{tab:qm9}
\begin{tabular}{@{}lccccccc@{}}
\toprule
Property & R-GIN+FA         & R-GAT+FA         & R-GatedGNN+FA    & GNN-FiLM         & SPN                      & $\ouracro$-GIN           & $\rate_1\ouracro$-GIN    \\ \midrule
mu       & 2.54\small±0.09  & 2.73\small±0.07  & 3.53\small±0.13  & 2.38\small±0.13  & 2.32\small±0.28          & \textbf{1.93\small±0.06} & 2.00\small±0.05          \\
alpha    & 2.28\small±0.04  & 2.32\small±0.16  & 2.72\small±0.12  & 3.75\small±0.11  & 1.77\small±0.09          & \textbf{1.63\small±0.03} & \textbf{1.63\small±0.05} \\
HOMO     & 1.26\small±0.02  & 1.43\small±0.02  & 1.45\small±0.04  & 1.22\small±0.07  & 1.26\small±0.09          & \textbf{1.16\small±0.01} & \textbf{1.17\small±0.02} \\
LUMO     & 1.34\small±0.04  & 1.41\small±0.03  & 1.63\small±0.06  & 1.30\small±0.05  & 1.19\small±0.05          & \textbf{1.13\small±0.02} & \textbf{1.15\small±0.02} \\
gap      & 1.96\small±0.04  & 2.08\small±0.05  & 2.30\small±0.05  & 1.96\small±0.06  & 1.89\small±0.11          & \textbf{1.74\small±0.02} & \textbf{1.74\small±0.03} \\
R2       & 12.61\small±0.37 & 15.76\small±1.17 & 14.33\small±0.47 & 15.59\small±1.38 & 10.66\small±0.40         & \textbf{9.39\small±0.13} & 9.94\small±0.07          \\
ZPVE     & 5.03\small±0.36  & 5.98\small±0.43  & 5.24\small±0.30  & 11.00\small±0.74 & \textbf{2.77\small±0.17} & \textbf{2.73\small±0.19} & \textbf{2.90\small±0.30} \\
U0       & 2.21\small±0.12  & 2.19\small±0.25  & 3.35\small±1.68  & 5.43\small±0.96  & 1.12\small±0.13          & \textbf{1.01\small±0.09} & \textbf{1.00\small±0.07} \\
U        & 2.32\small±0.18  & 2.11\small±0.10  & 2.49\small±0.34  & 5.95\small±0.46  & \textbf{1.03\small±0.09} & \textbf{0.99\small±0.08} & \textbf{0.97\small±0.04} \\
H        & 2.26\small±0.19  & 2.27\small±0.29  & 2.31\small±0.15  & 5.59\small±0.57  & \textbf{1.05\small±0.04} & \textbf{1.06\small±0.09} & \textbf{1.02\small±0.09} \\
G        & 2.04\small±0.24  & 2.07\small±0.07  & 2.17\small±0.29  & 5.17\small±1.13  & \textbf{0.97\small±0.06} & 1.06\small±0.14          & \textbf{1.01\small±0.05} \\
Cv       & 1.86\small±0.03  & 2.03\small±0.14  & 2.25\small±0.20  & 3.46\small±0.21  & 1.36\small±0.06          & \textbf{1.24\small±0.02} & \textbf{1.25\small±0.03} \\
Omega    & 0.80\small±0.04  & 0.73\small±0.04  & 0.87\small±0.09  & 0.98\small±0.06  & 0.57\small±0.04          & \textbf{0.55\small±0.01} & 0.60\small±0.03          \\ \bottomrule
\end{tabular}
\end{table*}

\paragraph{Experimental details.} Our experimental setup is based on \citet{brockschmidt2020gnn} and uses the same fixed data splits. We use the overall-best-performing SPN parameterisation with a `max distance' of $k_\text{max}=10$, referring to the $k$-hop neighbourhood aggregated over at each layer. This allows every node to interact with every other node at each layer when applied to a small-graph dataset like \texttt{QM9}, amounting to a dense static rewiring.
$\ouracro$-GIN and SPN models use a {\bf parameter budget} of \mytilde800,000, use 8 layers and train for 300 epochs; results are averaged over three runs. Neither SPN or $\ouracro$-GIN use relational edge features (denoted `R-'; \citet{schlichtkrull2018modeling, brockschmidt2020gnn}) as its impact is minimal (see Appendix \ref{appendix:experiments}). Other results are reproduced from their respective works \citep{brockschmidt2020gnn, alon2020bottleneck}; several of these include a final fully adjacent layer (+FA) which we include rather than the base models as they afford improved performance overall.

\textbf{Discussion.} $\ouracro$ demonstrates improvement over the classical and multi-hop MPNN benchmarks, beating or matching the next-best model, SPN, for 12 out of 13 regression targets. We note that, overall, the best average performance across targets is achieved by $\ouracro$ without delay ($\rate=\infty$).
This is as we might expect, as $L$-layer models with `slow' information flow such as classical GCNs and $\rate_1\ouracro$ cannot guarantee direct interaction between all node pairs on graphs with maximum diameter $>L$.

\subsection{Validating robustness}
In this section we demonstrate the robustness properties, rather than raw performance, of $\rate\ouracro$ with increasing network depth for long-range tasks.

\subsubsection{RingTransfer}
\begin{figure}[ht]
\begin{center}
\centerline{\includegraphics[width=0.8\columnwidth]{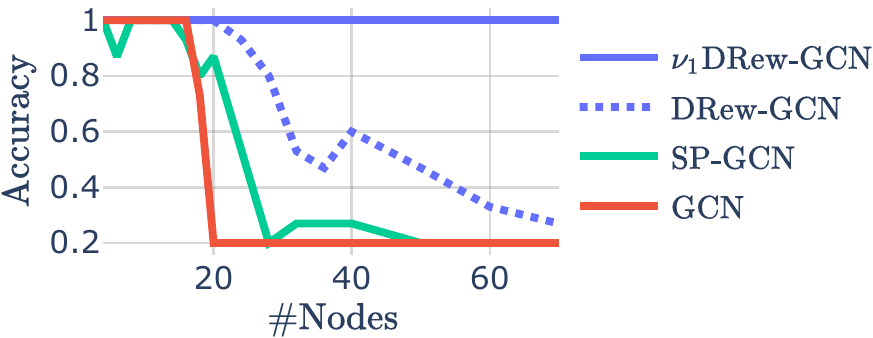}}
\caption{Performance on \texttt{RingTransfer} task for models with varying $k,L$. Accuracy of 0.2 corresponds to a random guess.}
\label{Fig:ringtransfer}
\end{center}
\vspace{-0.3in}
\end{figure}

\texttt{RingTransfer} is a synthetic task for empirically validating the ability of a GNN to capture {\bf long-range} node dependencies \cite{bodnar2021weisfeilercell}. The dataset consists of $N$ ring graphs (chordless cycles) of length $k$. Each graph has a single \textbf{source} node and a single \textbf{target} node that are always $\lfloor\frac{k}{2}\rfloor$ hops apart. Source node features are one-hot class label vectors of length $C$; all other nodes features are uniform. The task is for the \textbf{target} node to output the correct class label at the source. %
We compare $(\rate)\ouracro$-GCN againt a GCN and SP-GCN, %
an instance of the SPN framework \cite{abboud2022shortest}. %
Results are given in Figure \ref{Fig:ringtransfer} where the number %
of layers $L %
=\lfloor\frac{k}{2}\rfloor$ is the minimum required depth for source-target interaction. %

\vspace{-2mm}

\paragraph{Discussion.} \texttt{RingTransfer} demonstrates the power of $\rate\ouracro$ in mitigating MPNN performance issues brought on by increased depth. While the classical GCN fails after fewer than 10 layers, $\rate_1\ouracro$ achieves strong performance for 30 or more. These results also allow us to directly assess the impact of delay. The `full-delay' $\rate_1\ouracro$-GCN consistently achieves perfect accuracy for 30+ layers with no drop in performance. We can attribute this to the direct interaction between the target and delayed source node.
SP-GCN, however, with its static rewiring and dense computational graph, improves on the classical GCN, likely due to increased expressivity, but still fails at much shallower $L$ than $\rate\ouracro$, \emph{with or without delay}.

\subsubsection{Layerwise performance on Peptides-func}

\begin{figure}[ht]
\begin{center}
\centerline{\includegraphics[width=\columnwidth]{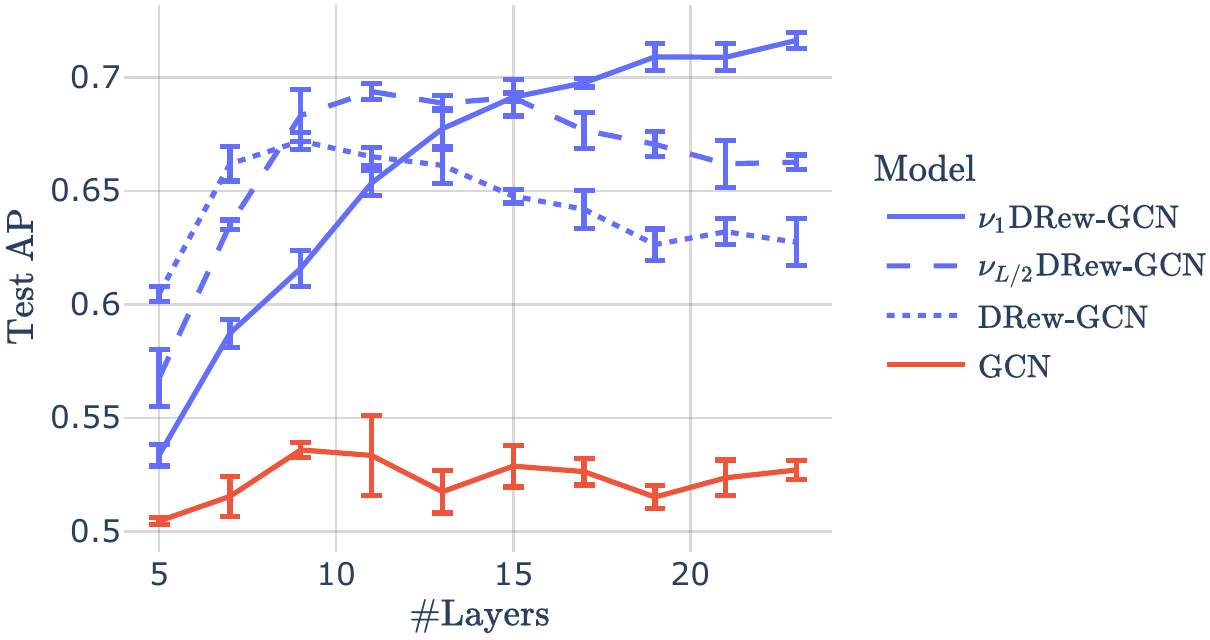}}
\vspace{-0.13in} %
\caption{Comparing three parameterizations of $\rate\ouracro$ plus a classical residual GCN on \texttt{Peptides-func} over varying $L$.}
\label{Fig:pept-func_d=64}
\end{center}
\vspace{-0.3in}
\end{figure}

 In this section we strip back our experiments on \texttt{Peptides-func} to demonstrate the robustness of  $\rate\ouracro$ for increasing network depth $L$, as well as the impact of $\rate$, the parameter denoting rate of information flow during message passing. For these experiments we fix the hidden dimension to 64.
 In Figure \ref{Fig:pept-func_d=64} we plot model performance against $L$ for three different parameterisations of $\rate\ouracro$-GCN: the non-delay version $\rate=\infty$ (which reduces to $\ouracro$), the full-delay version $\rate=1$, and a midpoint, where we set $\rate=L/2$.

\paragraph{Discussion.} Figure \ref{Fig:pept-func_d=64} demonstrates a crucial contribution of our framework: the ability to tune $\rate$ to suit the task.
 It is evident that using $\rate\ouracro$ with delay ensures more robust training when using a deeper architecture; in fact, \textbf{the more delay used} (i.e. the lower the value of $\rate$), \textbf{the better the performance} for large $L$, whereas using less delay (high $\rate$) ensures faster filling of the computational graph and greater density of connections after fewer layers. This means that, when using lower $L$, non-delay $\ouracro$ often performs better, especially when combined with PE.
 Conversely, more delay `slows down' the
 densification of node connections, yielding stronger long-term performance with $L$.
 Figure \ref{Fig:pept-func_d=64} demonstrates this with a long-range task:
 $\rate_1\ouracro$ consistently improves with more layers.

\section{Related Work}
Various methods have been developed to improve learning on graphs and %
avoid issues such as over-squashing. %
Many of these are broadly classified as `graph rewiring' methods; one such family involves sampling of nodes and/or edges of the graph based on some sort of performance or topological metric. %
Examples include sparsification \cite{hamilton2017inductive}, node (edge) dropout \cite{rong2019dropedge, papp2021dropgnn}, rewiring based on graph diffusion \cite{klicpera2019diffusion}, Cayley graphs \cite{deac2022expander}, commute times, spectral gap or Ricci curvature to combat over-squashing \citep{arnaiz2022diffwire, topping2021understanding, black2023understanding}.
Most of these methods remove supposedly irrelevant elements from the graph to make it more amenable to analysis, though many methods also add elements to increase connectivity. This might be a global node \cite{battaglia2016interaction, gilmer2017neural}, or global layer, such as positional/structural encoding \cite{dwivedi2020benchmarking, dwivedi2021graph, rampavsek2022recipe, wang2022equivariant}, or adding a fully adjacent layer after message passing \cite{alon2020bottleneck}. It may also take the form of multiple-hop rewiring, in which aggregations occur over nodes at $>$1-hop distance at each layer; we distinguish these into `local' graph rewiring, also known as multi-hop MPNNs \cite{abboud2022shortest, abu2019mixhop, abu2020n, nikolentzos2020k, zhang2021nested}, and `global' methods such as graph Transformers \cite{dwivedi2022long, kreuzer2021rethinking, rampavsek2022recipe, ying2021transformers, yun2019graph}, which %
fully connect the input graph. %

Unlike all of these methods, the rewiring used in $\ouracro$ is layer-dependent, i.e. it is adaptive rather than static. Our method is also unique in its ability to control the rate of information flow by tuning the delay parameter $\rate$. Our use of delay is loosely inspired by delay differential equations (DDEs), which have also inspired architectures which leverage delay in the wider deep learning space \citep{anumasa2021delay, zhu2021neural, zhu2022neural} based on neural ordinary differential equations \cite{chen2018neural}, but we know of no DDE-inspired works in the graph machine learning space.
The idea of accessing previous node states resonates with \citet{xu2018representation} and \citet{strathmann2021persistent}.  \citet{faber2022asynchronous} use a form of delay to create node identifiers as a means to allow nodes to ignore irrelevant messages from their neighbours, but all of these works bear little relation to $\ouracro$, which treats dynamic rewiring and delay from the perspective of distance on graphs.

\section{Conclusion and Future Work}
We have introduced an adaptive MPNN framework based on a layer-dependent dynamic rewiring that can be adapted to any MPNN. We have also proposed a delay mechanism permitting local skip-connections among different nodes based on their mutual distance. Investigating the expressive power of this framework represents a promising future avenue to compare static and dynamic rewiring approaches, as well as the impact of distance-aware skip-connections.

\vspace{-1mm}

\paragraph{Limitations.}
Our framework is expected to be useful for tasks with long-range interactions %
or, more generally, when one requires very deep GNN models, as confirmed by our experiments. %
Accordingly, we do not expect our framework to be advantageous when applied to (for example) homophilic node classification tasks where shallow GNNs acting as low-pass filters are sufficient to perform strongly. This may partly explain why $\ouracro$ is outperformed by GraphGPS for \voc\ (see Table \ref{tab:all_results}), as %
this dataset %
presents %
an image segmentation task which likely displays a reasonable degree of homophily. As a result, the extent to which long-range interactions are truly present is uncertain.

\paragraph{Acknowledgements.}
We are grateful for anonymous reviewer feedback. We acknowledge the use of the University of Oxford Advanced Research Computing (ARC) facility in carrying out this work \cite{richards2021arc}
, as well as the
JADE
HPC facility.
B.G. acknowledges support from the EPSRC Centre for Doctoral Training in AIMS (EP/S024050/1).
X.D. acknowledges support from the Oxford-Man Institute of Quantitative
Finance and the EPSRC (EP/T023333/1).
M.B. is supported in-part by ERC Consolidator Grant No. 274228 (LEMAN) and Intel AI Grant.

\newpage
\bibliography{references}
\bibliographystyle{icml2023}

\newpage
\appendix
\onecolumn

\section{Time and Space Complexity}
\label{appendix:complexity}
\paragraph{Time complexity.} $\ouracro$ relies on the shortest path and therefore requires up-to $k$-hop adjacency information for layer $k$. This can be pre-computed in worst-case time $\mathcal{O}(|V||E|)$ using the same breadth-first search method of \citet{abboud2022shortest}, but once computed it can be re-used for all runs.

In the worst case --- when $\ell$ is greater than the max diameter of the graph --- $\ouracro$ performs aggregation over $\mathcal{O}(V^2)$ elements, i.e. all node pairs. However, each $k$-neighbourhood aggregation can be computed in parallel, so this is not a serious concern in practice, and as $\ouracro$ gradually builds the computational graph at each layer, it is faster than comparable multi-hop MPNNs or Transformers that aggregate over the entire graph, or a fixed hop distance, at every layer.

\paragraph{Space complexity.} As we use delayed representations of nodes, we must store $\ell$ representations of the node features at layer $\ell$ for a given mini-batch; i.e. linear scaling in network depth $L$. This has not proved to be a bottleneck in practice, and could be addressed if need be by reducing batch size.

\section{Further Experimental Details}
\label{appendix:experiments}
In this section we provide further details about our experiments, as well as further results on \texttt{Peptides-func}.

\subsection{Hardware}
All experiments were run on server nodes using a single GPU. A mixture of P100, V100, A100, Titan V and RTX GPUs were used, as well as a mixture of Broadwell, Haswell and Cacade Lake CPUs.

\subsection{Long range graph benchmark}

All results in Tables \ref{tab:mpnn_vs_drew_all_lrgb} and \ref{tab:all_results} use similar experimental setups and identical data train--val--test splits to their classical MPNN counterparts in \citet{dwivedi2022long}. We use a fixed parameter budget of \mytilde500,000, which is controlled for by appropriate tuning of hidden dimension when using different network depths $L$.
Significant hyperparameter differences for experiments are given in Table \ref{tab:all_lrgb_hyp}; other experimental details are itemised below:
\begin{itemize}
    \item For all experiments we train for 300 epochs or until convergence and average results over three seeds.
    \item All results use the AdamW optimizer \cite{kingma2017adam, loshchilov2017decoupled} with base learning rate $\code{lr=0.001}$ (except \texttt{PascalVOC-SP}, which uses $\code{0.0005}$), $\code{lr\_decay=0.1}$,
    $\code{min\_lr=1e-5}$,
    $\code{momentum=0.9}$,
    and a reduce-on-plateau scheduler with $\code{reduce\_factor=0.5}$,  $\code{schedule\_patience=10}$ ($\code{20}$ for \peptides).
    \item All \peptides, \pcqm\ and \voc\ experiments use batch sizes of 128, 256 and 32 respectively.
    \item All experiments use batch normalisation with $\code{eps=1e-5}$, $\code{momentum=0.1}$ and post-layer $L_2$ normalisation.
    \item \peptides\ and \pcqm\ use the `Atom' node encoder \cite{hu2020open, hu2021ogb}, whereas \voc\ uses a node encoder which is a single linear layer with a fixed output dimension of 14.
    \item None of the experiments in Table \ref{tab:all_results} use edge encoding or edge features.
    \item Superpixel nodes in \voc\ are extracted using a SLIC compactness of 30 for the SLIC algorithm \cite{achanta2012slic}.
    \item We do not use dropout.
    \item All Laplacian PE uses hidden dimension of 16 and 2 layers.
    \item For \pcqm, all experiments (except for DIGL) use convex combination with equal weights for aggregation; i.e. each of $M$ channel-mixing matrices per $k$-neighbourhood aggregation is multiplied by $1/M$. Other tasks do not use any matrix weights.
    \item Experiments for MixHop-GCN, a a multi-hop MPNN, are parameterised by $\max P$,  where integer powers of the adjacency matrix are aggregated up to $\max P$, with equal-size weight matrices per adjacency power. The hyperparameters in Table \ref{tab:all_lrgb_hyp} were determiend by best performance after hyperparameter search over $\max P$ and network depth $L$.
    \item We perform DIGL rewiring using the default settings from the Graph Diffusion Convolution transform from \code{torch\_geometric.transforms} using PPR diffusion with $\alpha=0.2$ and threshold sparsification with average degree $d_{\text{avg}}$ given in Table \ref{tab:all_lrgb_hyp}.
    \item All DIGL+MPNN runs use GCN as the base MPNN except for \texttt{PascalVOC-SP} which uses GatedGCN instead for fair comparison, as other classical MPNNs perform poorly on this task
    \item \func\ results in Figures \ref{Fig:pept-func_d=64}, \ref{Fig:func_ablation} and \ref{Fig:func_lappe_ablation} use the same experimental setup as described above.
    \item The reported results for GatedGCN+PE in Table \ref{tab:all_results} use LapPE for \texttt{PascalVOC-SP}and RWSE for all other tasks.

\end{itemize}

\begin{table}[h]
\caption{Parameter counts (\#Ps) and significant hyperparameters (HPs) for for all DIGL, MixHop-GCN and $(\rate)\ouracro$-MPNN experiments in Table \ref{tab:all_results}. Hyperparameterisation details for reproduced results are available in their respective works.}
\label{tab:all_lrgb_hyp}
\begin{tabular}{@{}llrlrlrlr@{}}
\toprule
\textbf{Model}      & \multicolumn{2}{c}{\texttt{Peptides-func}}                                 & \multicolumn{2}{c}{\texttt{Peptides-struct}}                              & \multicolumn{2}{c}{\texttt{PCQM-Contact}}                                 & \multicolumn{2}{c}{\texttt{PascalVOC-SP}}                                  \\
                    & \#Ps & HPs                                                                 & \#Ps & HPs                                                                & \#Ps & HPs                                                                & \#Ps & HPs                                                                 \\ \midrule
DIGL+MPNN           & 499k & \begin{tabular}[c]{@{}r@{}}$d_{\text{avg}}=6$\\ $L=13$\end{tabular} & 496k & \begin{tabular}[c]{@{}r@{}}$d_{\text{avg}}=6$\\ $L=5$\end{tabular} & 497k & \begin{tabular}[c]{@{}r@{}}$d_{\text{avg}}=2$\\ $L=5$\end{tabular} & 502k & \begin{tabular}[c]{@{}r@{}}$d_{\text{avg}}=14$\\ $L=8$\end{tabular} \\ \midrule
DIGL+MPNLapPE     & 493k & \begin{tabular}[c]{@{}r@{}}$d_{\text{avg}}=6$\\ $L=5$\end{tabular}  & 496k & \begin{tabular}[c]{@{}r@{}}$d_{\text{avg}}=6$\\ $L=7$\end{tabular} & 495k & \begin{tabular}[c]{@{}r@{}}$d_{\text{avg}}=2$\\ $L=5$\end{tabular} & 502k & \begin{tabular}[c]{@{}r@{}}$d_{\text{avg}}=14$\\ $L=8$\end{tabular} \\ \midrule
MixHop-GCN          & 513k & \begin{tabular}[c]{@{}r@{}}$\max P=5$\\ $L=17$\end{tabular}         & 510k & \begin{tabular}[c]{@{}r@{}}$\max P=7$\\ $L=17$\end{tabular}        & 523k & \begin{tabular}[c]{@{}r@{}}$\max P=3$\\ $L=5$\end{tabular}         & 511k & \begin{tabular}[c]{@{}r@{}}$\max P=5$\\ $L=8$\end{tabular}          \\ \midrule
MixHop-GCN+LapPE    & 518k & \begin{tabular}[c]{@{}r@{}}$\max P=7$\\ $L=14$\end{tabular}         & 490k & \begin{tabular}[c]{@{}r@{}}$\max P=7$\\ $L=11$\end{tabular}        & 521k & \begin{tabular}[c]{@{}r@{}}$\max P=3$\\ $L=5$\end{tabular}         & 512k & \begin{tabular}[c]{@{}r@{}}$\max P=5$\\ $L=8$\end{tabular}          \\ \midrule
DRew-GCN            & 518k & \begin{tabular}[c]{@{}r@{}}$\nu=1$\\ $L=23$\end{tabular}            & 498k & \begin{tabular}[c]{@{}r@{}}$\nu=\infty$\\ $L=13$\end{tabular}      & 515k & \begin{tabular}[c]{@{}r@{}}$\nu=\infty$\\ $L=20$\end{tabular}      & 498k & \begin{tabular}[c]{@{}r@{}}$\nu=1$\\ $L=8$\end{tabular}             \\ \midrule
DRew-GCN+LapPE      & 502k & \begin{tabular}[c]{@{}r@{}}$\nu=\infty$\\ $L=7$\end{tabular}        & 495k & \begin{tabular}[c]{@{}r@{}}$\nu=\infty$\\ $L=5$\end{tabular}       & 498k & \begin{tabular}[c]{@{}r@{}}$\nu=\infty$\\ $L=10$\end{tabular}      & 498k & \begin{tabular}[c]{@{}r@{}}$\nu=1$\\ $L=8$\end{tabular}             \\ \midrule
DRew-GIN            & 514k & \begin{tabular}[c]{@{}r@{}}$\nu=1$\\ $L=17$\end{tabular}            & 505k & \begin{tabular}[c]{@{}r@{}}$\nu=\infty$\\ $L=15$\end{tabular}      & 507k & \begin{tabular}[c]{@{}r@{}}$\nu=\infty$\\ $L=20$\end{tabular}      & 506k & \begin{tabular}[c]{@{}r@{}}$\nu=1$\\ $L=8$\end{tabular}             \\ \midrule
DRew-GIN+LapPE      & 502k & \begin{tabular}[c]{@{}r@{}}$\nu=1$\\ $L=15$\end{tabular}            & 497k & \begin{tabular}[c]{@{}r@{}}$\nu=\infty$\\ $L=5$\end{tabular}       & 506k & \begin{tabular}[c]{@{}r@{}}$\nu=\infty$\\ $L=10$\end{tabular}      & 506k & \begin{tabular}[c]{@{}r@{}}$\nu=1$\\ $L=8$\end{tabular}             \\ \midrule
DRew-GatedGCN       & 495k & \begin{tabular}[c]{@{}r@{}}$\nu=1$\\ $L=17$\end{tabular}            & 497k & \begin{tabular}[c]{@{}r@{}}$\nu=\infty$\\ $L=13$\end{tabular}      & 506k & \begin{tabular}[c]{@{}r@{}}$\nu=\infty$\\ $L=20$\end{tabular}      & 502k & \begin{tabular}[c]{@{}r@{}}$\nu=1$\\ $L=8$\end{tabular}             \\ \midrule
DRew-GatedGCN+LapPE & 495k & \begin{tabular}[c]{@{}r@{}}$\nu=\infty$\\ $L=7$\end{tabular}        & 494k & \begin{tabular}[c]{@{}r@{}}$\nu=\infty$\\ $L=5$\end{tabular}       & 494k & \begin{tabular}[c]{@{}r@{}}$\nu=\infty$\\ $L=10$\end{tabular}      & 502k & \begin{tabular}[c]{@{}r@{}}$\nu=1$\\ $L=8$\end{tabular}             \\ \bottomrule
\end{tabular}
\end{table}

\subsection{\texttt{QM9}}
Performance experiments use a fixed parameter budget of 800k, controlled for by appropriate tuning of the hidden dimension when using different network depth $L$. We mostly follow the experimental setup of \cite{abboud2022shortest}, using a fixed learning rate of 0.001, mean pooling and no dropout. We use batch normalization, train for 300 epochs averaging over 3 runs, and use data splits from \cite{brockschmidt2020gnn}.

Many of the benchmarks we compare against in Table \ref{tab:qm9} are Relational MPNNs (R-MPNN; \citet{schlichtkrull2018modeling, brockschmidt2020gnn}) which incorporate edge labels by assigning separate weights for each edge type in the 1-hop neighbourhood, and aggregating over each type separately.
For our SPN and $\ouracro$-GIN experiments, however, we do not incorporate edge features, as (a) $\ouracro$ demonstrates strong performance even without this information, and (b) we expect the improvement obtained through using R-GIN to be slight given that over 92\% of all graph edges in \texttt{QM9} are of only one type.

\subsection{\texttt{RingTransfer}}
For synthetic \texttt{RingTransfer} \cite{bodnar2021weisfeilercell} experiments we use a dataset of size $N=2000$ with $C=5$ classes and a corresponding node feature dimension. GCN and SP-GCN runs use a hidden dimension of 256, and for fair comparison $\ouracro$-GCN runs use a varying hidden dimension to ensure the same parameter count as GCN/SP-GCN for each ring length $k$ (and therefore model depth $L$). All runs use batch normalization, no dropout, and Adam optimization with learning rate 0.01. We train for 50 epochs and average over three experiments, using the accuracy of predicting the source node label from the readout of the source node representation as a metric. We use an 80:10:10 split for train, test and validation.

\paragraph{SP-GCN}
We define SP-GCN as an instance of the SP-MPNN framework \cite{abboud2022shortest}:
\begin{equation}
\label{Eq:alphaGCN}
   h^{(\ell+1)}_i = h^{(\ell)}_i + \sigma \left( \sum_{k=1}^{k_\text{max}} \sum_{j \in \mathcal{N}_k} \alpha_k^{(\ell)} \mathbf{W}^{(\ell)} \gamma^k_{ij} h^{(\ell)}_j \right),
\end{equation}
where $k_\text{max}$ is the max distance parameter that determines the number of hops to aggregate over at each layer and $\boldsymbol{\alpha}^{(\ell)} \subset \R^{k_\text{max}}$ are learned weight vectors, $\sum^{k_\text{max}}_k \alpha_k^{(\ell)}=1$. $\gamma^k_{ij}$ is as in Eq.~\eqref{eq:gamma}.

\subsection{Ablation on \texttt{Peptides-func}}
\begin{figure}[ht]
\begin{center}
\centerline{\includegraphics[width=0.8\linewidth]{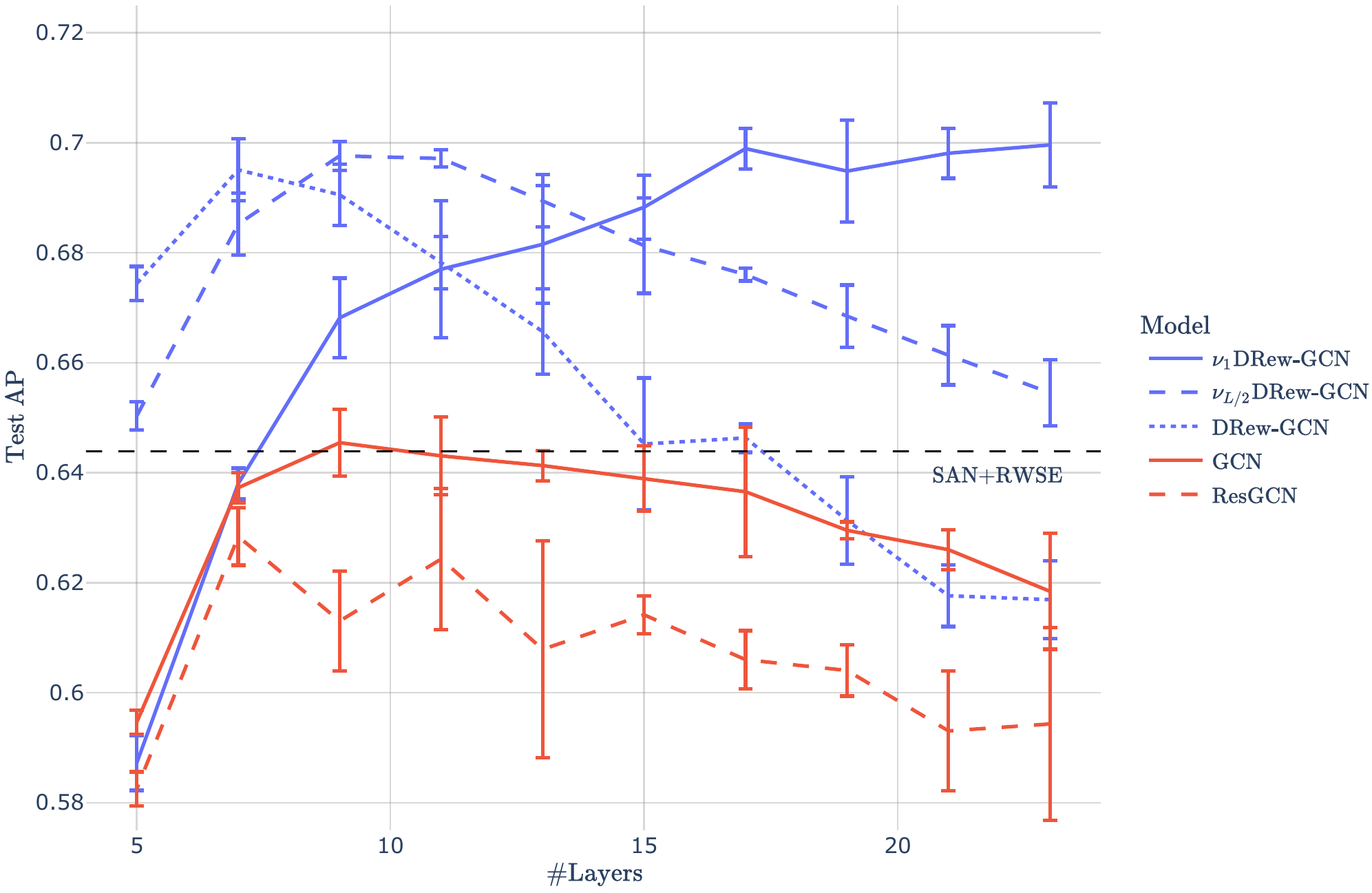}}
\caption{\func\ experiments over varying $L$ with fixed 500k parameter count using \textbf{no} positional encoding.}
\label{Fig:func_ablation}
\end{center}
\end{figure}

\begin{figure}[ht]
\begin{center}
\centerline{\includegraphics[width=0.8\linewidth]{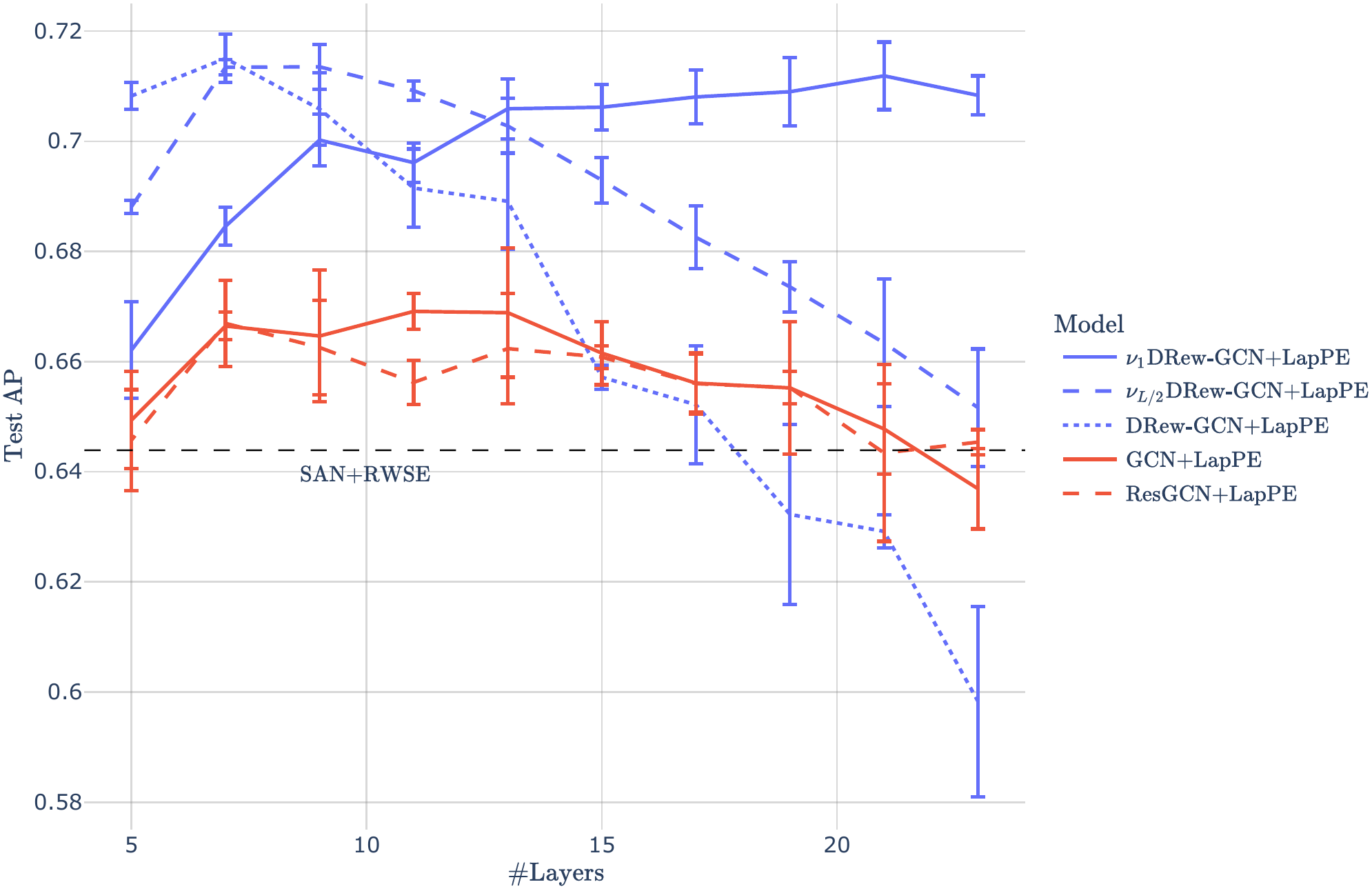}}
\caption{\func\ experiments over varying $L$ with fixed 500k parameter count using \textbf{Laplacian} positional encoding.}
\label{Fig:func_lappe_ablation}
\end{center}
\end{figure}

In this section we provide additional experimental results on \func\ using our 500k parameter budget setup from Section \ref{sec:lrgb}. We train a number of varying-depth GCN, residual GCN and $\rate\ouracro$-GCN models with three parameterizations of $\rate$: $1, L/2$ and $\infty$. We provide separate results with and without Laplacian PE \cite{dwivedi2020benchmarking} in Figures \ref{Fig:func_lappe_ablation} and \ref{Fig:func_ablation} respectively. For reference, on both figures we also mark the best-performing Transformer, SAN with random walk encoding \citep{kreuzer2021rethinking, dwivedi2021graph}, reproduced from \citet{dwivedi2022long} and denoted with a dashed black line.

\paragraph{Discussion}
From Figures \ref{Fig:func_ablation} and \ref{Fig:func_lappe_ablation}  we can see that more delay leads to stronger performance at greater network depths, even as the hidden dimension decreases severely. We see that low $L$, low/no delay $\rate\ouracro$ and high $L$, high delay $\rate\ouracro$ outperform GCNs and the best-performing Transformer, \emph{with or without positional encoding}.

We note that the poor performance of low-delay $\rate\ouracro$ at high $L$ and vice-versa is as expected. As \func\ is a long-range task, strong performance requires interaction between distant nodes. Though it uses more powerful multi-hop aggregations, $\rate_1\ouracro$ maintains the same `rate' of information flow as classical MPNNs, i.e. $r$-distant nodes are able to interact from the $r$th layer; therefore small $L$ does not give $\rate_1\ouracro$ sufficient layers to enable long-range interactions, and performance is comparable to classical MPNNs. As our framework is more robust, however, $\rate_1\ouracro$ continues to increase in performance as $L$ increases --- as we would hope --- while the classical MPNNs GCN and ResGCN succumb to the usual issues that affect MPNNs, and degrade.

Conversely, the low delay parameterizations, $\{\nu_{L/2}, \nu_\infty\}\ouracro$-GCN, perform strongly for low $L$ and worsen as the network becomes very deep. Again, this is expected: low delay affords fast (or instantaneous) communication between distant nodes after sufficient layers, and therefore has a rate of information flow that is faster than $\rate_1\ouracro$ or classical MPNNs (though still slower than multi-hop MPNNs or Transformers). This means that, for a long-range task such as \func, performance is stronger for fewer layers, once long-range interactions have been facilitated but \emph{before} the computational graph becomes too dense, causing performance drop. These experiments further demonstrate the impact and usefulness of the delay parameter as a means of tuning the model to suit the task.

Referring to Figure \ref{Fig:func_lappe_ablation}, we note that the addition of PE exacerbates the behaviours described above, and indeed accelerates the rate of information flow by preceding the message-passing with a global layer.

\paragraph{Positional encoding.}
As a final point, we consider the overall impact of PE, particularly Laplacian PE, on this task. We posit that, for \texttt{Peptides}, the characterization of Transformers as the strongest long-range models \cite{dwivedi2022long} is due primarily to PE, rather than the Transformer architecture. As evidence we point to the performance of vanilla GCN, the simplest MPNN, on \texttt{func} when LapPE is used; it outperforms SAN with only five layers.
We reiterate that $\rate\ouracro$ outperforms MPNN and Transformer+PE benchmarks with or without using PE itself.

\end{document}